\documentclass[a4paper]{article}
\usepackage{acl2014}
\usepackage[english]{babel}
\usepackage[utf8]{inputenc}
\usepackage{amsmath}
\usepackage{graphicx}
\usepackage[colorinlistoftodos]{todonotes}
\usepackage{subcaption}
\usepackage[export]{adjustbox}
\usepackage{amssymb}
\usepackage{algorithm}
\usepackage{algorithmicx}
\usepackage{algpseudocode}
\usepackage{multirow}
\usepackage{tikz}
\usetikzlibrary{trees}
\DeclareMathOperator*{\argmin}{arg\,min}

\title{Navigating the Semantic Horizon\\
using Relative Neighborhood Graphs}

\author{
Amaru Cuba Gyllensten and Magnus Sahlgren\\
Gavagai\\
Bondegatan 21\\
116 33 Stockholm\\
Sweden\\
\texttt{\{amaru|mange\}@gavagai.se}}

\begin{document}

\maketitle

\begin{abstract}
This paper is concerned with nearest neighbor search in distributional semantic models. A normal nearest neighbor search only returns a ranked list of neighbors, with no information about the structure or topology of the local neighborhood. This is a potentially serious shortcoming of the mode of querying a distributional semantic model, since a ranked list of neighbors may conflate several different senses. We argue that the topology of neighborhoods in semantic space provides important information about the different senses of terms, and that such topological structures can be used for word-sense induction. We also argue that the topology of the neighborhoods in semantic space can be used to determine the {\em semantic horizon} of a point, which we define as the set of neighbors that have a direct connection to the point. We introduce {\em relative neighborhood graphs} as method to uncover the topological properties of neighborhoods in semantic models. 
%Such relative neighborhood graphs are useful tools both for word-sense induction and for analyzing and comparing the structural properties of local neighborhoods in different semantic models. 
We also provide examples of relative neighborhood graphs for three well-known semantic models; the PMI model, the GloVe model, and the skipgram model.
\end{abstract}

\section{Introduction}

Nearest neighbor search is fundamental operation in data mining, in which we are interested in finding the closest points (to some given reference point). Formally, if we have a reference point $r$ and a set of other points $P$ in a metric space $M$ with some distance function $d$ (or similarity function $s$), the nearest neighbor search task is to find the point $p \in P$ that minimizes $d(p,r)$. In $k$-nearest neighbor search ($k$-NN), we want to find the $k$ closest points to some given reference point. Nearest neighbor search is a well-studied task, and in particular the complexity of the task (a linear search has a running time of $\mathcal{O}(Ni)$ where $N$ is the cardinality of $P$ and $i$ the complexity of the distance function $d$) has generated a lot of research \cite{Bentley:1975,Arya:1998,Indyk:1998}.
%; suggestions for reducing the complexity of linear nearest neighbor searches include using various types of space partitioning techniques like $k$-d trees \cite{Bentley:1975}, or various techniques for doing {\em approximate} nearest neighbor search \cite{Arya:1998}, of which one of the most well-known is locality-sensitive hashing \cite{Indyk:1998}.

\begin{figure}
\includegraphics[scale=0.3]{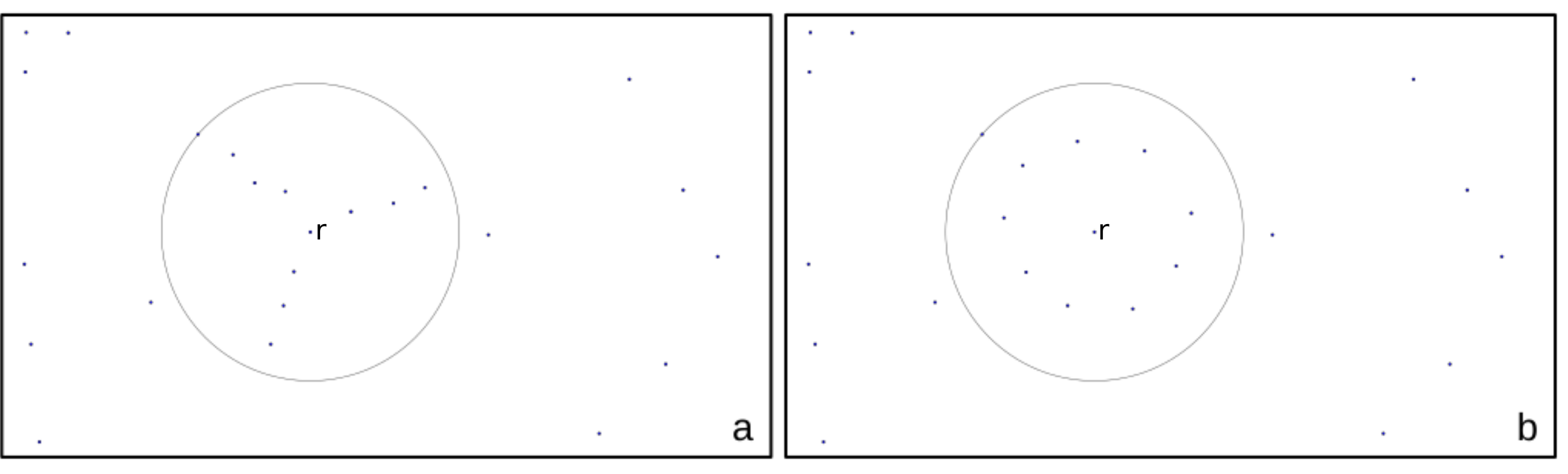}
\caption{Examples of neighborhoods with a clear branching structure ({\bf a}) and without ({\bf b}).}
\label{fig:twospaces}
\end{figure}

The problem we are concerned with in this paper is not the complexity of nearest neighbor search, but the question {\em how to identify the internal structure of neighborhoods defined by the nearest neighbors}. The problem with a normal $k$ nearest neighbor search is that the result (a sorted list of the $k$ nearest neighbors) does not say anything about the internal structure of the neighborhood. Consider spaces $a$ and $b$ in Figure \ref{fig:twospaces}. A nearest neighbor search for the reference point $r$ in these two spaces will generate the exact same result, despite the fact that the neighborhoods are very different with regards to their internal structure (the neighbors in space $a$ display a distinct branching structure, whereas the neighbors in space $b$ are distributed evenly across the space). Such structural properties of nearest neighborhoods can be very important.

We propose to use {\em relative neighborhood graphs} in order to identify the structural properties of nearest neighborhoods. The use of relative neighborhood graphs also provides a partial solution 
% * <magnus.sahlgren@gavagai.se> 2014-11-24T14:42:17.893Z:
% jag skriver "partial" eftersom vi löser problemet med att definiera k map hur många kluster vi upptäcker, men inte map hur många grannar vi populerar klustren med
% ^ <magnus.sahlgren@gavagai.se> 2015-01-02T15:47:53.239Z.
to the problem of finding a relevant $k$ for a given reference point. Again, consider the neighborhoods in Figure \ref{fig:twospaces}. The choice of $k=10$ in these spaces is completely arbitrary, and could be argued to be erroneous, since there are in fact 12 relevant neighbors in both spaces. Another way to approach the nearest neighbor search task is to use a radius around the reference point, so that only points within that radius are considered to be neighbors. 
%This is equivalent to defining a threshold $t$ and only counting points with $D(s,r) < t$ as neighbors to $r$. 
However, setting a global threshold $t$ seems just as arbitrary as setting a global number of neighbors $k$. Ideally, $t$ or $k$ should be determined based on the structural properties of the nearest neighborhood around the reference point. We refer to this factor
% * <magnus.sahlgren@gavagai.se> 2014-11-24T15:45:28.354Z:
% vetefan vad korrekt term är här - använder "factor" så länge
% ^ <magnus.sahlgren@gavagai.se> 2015-01-07T12:09:41.248Z.
as the {\em horizon} with respect to the reference point.

Although a relative neighborhood graph is a general method for defining and structuring nearest neighborhoods, we are in this paper primarily interested in its application to nearest neighbor searches in {\em distributional semantic models} that collect and represent co-occurrence statistics 
% of terms 
in high-dimensional vector spaces. The main operation in such models is nearest neighbor search, which is used for finding terms that have similar co-occurrence behavior. However, a ranked list of neighbors does not provide any information on whether the neighbors belong to several different senses. This problem has been misinterpreted as a shortcoming of the distributional representation \cite{Erk:2010}. However, as we will demonstrate in this paper, this is not a shortcoming of the distributional representation, but of the {\em mode of querying} the distributional model. We argue that information about the different usages (i.e.~senses) of a term is encoded in the structural properties of the nearest neighborhoods, and that a relative neighborhood graph is a viable tool for uncovering such structural properties. 

%In the following sections, we provide an overview of distributional semantics and its use of nearest neighbor search (Section \ref{sec:dsm}). We then review previous approaches to detecting polysemy and performing word sense induction in distributional semantic models (Section \ref{sec:wsi}), before introducing the use of relative neighborhood graphs and their application to distributional semantics (Section \ref{sec:rng}). We also discuss problems with evaluating nearest neighbor search methods qualitatively, and exemplify how relative neighborhood graphs can be used for word sense induction.

\section{Distributional Semantics and Nearest Neighbor Search}
\label{sec:dsm}

Collecting and comparing co-occurrence statistics for terms in language has become a standard approach for computational semantics, and is now commonly referred to as {\em distributional semantics}. There are many different types of models that can be used for this purpose, but their common objective is to represent terms as vectors that record (some function of) their distributional properties. The standard approach for generating such vectors is to collect distributional statistics in a {\em co-occurrence matrix} that records co-occurrence counts between terms and contexts. The co-occurrence matrix is then subject to various types of transformations, ranging from the application of simple frequency filters or association measures like pointwise mutual information, to matrix factorization or regression models.
% * <magnus.sahlgren@gavagai.se> 2014-12-05T16:31:41.288Z:
%  bara "regression" eller "regression models"?
% ^ <magnus.sahlgren@gavagai.se> 2015-01-02T15:48:09.310Z.
The resulting representations are referred to as {\em distributional vectors}, 
%% skippa resten?
and are typically dense with a dimensionality that is considerably lower than that of the original co-occurrence matrix. 

The distributional vectors are used to compute similarity between terms. There are many ways to compute similarity or distance between points in vector space; the cosine of the angle between vectors is often the preferred metric in distributional semantics because of its simplicity and because it normalizes for vector length.
%\footnote{$\frac{a\cdot b}{\|a\|\|b\|} = \frac{\sum_{i=1}^n a_i b_i}{\sqrt{\sum_{i=1}^n a^{2}_{i}} \sqrt{\sum_{i=1}^n b^{2}_{i}}}$}
Computing the similarity between distributional vectors using the cosine measure gives us a score ranging from $-1$ --- negatively collinear --- to $1$ --- positively collinear ---  taking the value 0 if the vectors are orthogonal.

We can thus use a distributional semantic model to quantify the similarity between any given terms. If the set of given terms is the {\em entire} set of terms in our model, we are in effect performing a nearest neighbor search. This is a particularly important operation in distributional semantics, since it answers the question "which other terms are similar to this one?", and this is a central question in semantics; lexica and thesauri are built with the main purpose of answering this question, and a nearest neighbor search in a distributional semantic model could therefore be seen as a compilation step in a distributional lexicon.

The result of a nearest neighbor search in a distributional semantic model is often presented as a list of (the top $k$) neighbors, sorted by descending similarity with the target term. Table~\ref{tab:nnlist} illustrates typical sorted nearest neighbor lists produced with three different kinds of distributional semantic models: a vanilla-flavored model based on (positive) Pointwise Mutual Information (PMI),\footnote{For observations $a$ and $b$, pmi($a,b$)$=\log \frac{p(a,b)}{p(a)p(b)}$. The probabilities are often replaced in distributional semantic models by co-occurrence counts of $a$ and $b$ and their respective frequency counts.} the skip-gram model \cite{Mikolov:2013}, and GloVe \cite{Pennington:2014}.

\begin{table}[h]
\centering
\caption{Sorted list of the nearest neighbors to ``suit'' in different distributional models. Different fonts represent different meanings of "suit."}
\label{tab:nnlist}
\begin{tabular}{|l|l|l|}
\hline
PMI & GloVe & skipgram \\
\hline
suits&suits&suits\\
dress&lawsuit&lawsuit\\
jacket&filed&countersuit\\
wearing&case&classaction\\
hat&wearing&doublebreasted\\
trousers&laiming&skintight\\
costume&lawsuits&necktie\\
shirt&alleging&wetsuit\\
pants&alleges&crossbone\\
lawsuit&classaction&lawsuits\\
\hline
\end{tabular}
\end{table}

In the vanilla-flavored model, the distributional vector of a word is given by its (positive) PMI with regards to all other words that have occurred within a context window of 2 words to the left and 2 words to the right. That is, a vector for a word $a$ corresponds to the information an observation of $a$ gives when predicting surrounding words. {\em Positive} PMI means that negative values are discarded, and only positive PMI values are retained. The cosine similarity of two distributional vectors thus gives a measure of how similar the information gained
by observing the corresponding words are. As a way to speed up later computations we apply a Gaussian random projection to reduce the dimensionality down to 2000.

GloVe on the other hand tries to find distributional vectors such that their dot product approximates their log probability of co-occurring is motivated by the fact that the logarithm of ratios equals the difference of logarithms, which makes the vector differences meaningful in that they encode (logarithms of) ratios of probabilities. Reframed as a weighted least squares problem, where rare co-occurrences are weighted down, it can be solved by standard methods. The performance is comparable to the skip-gram model, and it performs particularly well on word-analogy tasks \cite{Pennington:2014}.

The objective of the skipgram model is to maximize the probability of observing all context-word pairs given that the probability of one observation of a word $c$ in the context of $t$ is given by $\frac{\exp{(w_c^\top v_t)}}{\sum_{l \in V} \exp(u_l^\top v_t)}$ where  $v_a$ and $u_a$ denotes the "input" and "output" vectors of the word $a$, and $V$ is the vocabulary. The embeddings are found using stochastic gradient descent and hierarchical softmax combined with negative sampling and subsampling. Exactly how these methods compose is still unclear, and puts into question what the underlying model actually is \cite{Levy:2014}. Regardless, the skip-gram model delivers state of the art performance on a multitude of tasks, with very low-dimensional vectors \cite{Baroni:2014}.

Table~\ref{tab:nnlist} lists the 10 nearest neighbors to \emph{suit} in three different distributional semantic models using the entire Wikipedia as data.\footnote{We use a Wikipedia dump from 2010 as data in this and following experiments.} 
As can be expected, there are both similarities and dissimilarities between these neighborhoods; "suits" and "lawsuit" occur among the 10 nearest neighbors to "suit" in all three models, whereas other terms are specific for one particular model. Yet all three models feature neighbors of  "suit" that represent different senses:  the way "suit" is not related to "jacket" in the same way it is related to "lawsuit".   
%As can be expected, there are both similarities and dissimilarities between these neighborhoods; "suits" and "lawsuit" occur among the 10 nearest neighbors to "suit" in all three models, whereas other terms are specific for one particular model. The PMI model features terms like "dress", "jacket" and "wearing", the GloVe model features terms like "filed", "case" and "claiming", while the skipgram model includes terms like "countersuit", "classaction" and "doublebreasted". All three models feature neighbors that represent two different usages of "suit": the {\em law}-related sense ("lawsuit") and the {\em clothes}-related sense ("dress", "wearing", "doublebreasted"). The skipgram model also features a {\em manga}-related sense of "suit" in the neighbor "crossbone," which refers to the mange series "Mobile Suit Crossbone Gundam." However, these distinction are not discernible by merely looking at the list of nearest neighbors; the only information it provides is the ranking of the nearest neighbors in descending order of similarity. 

It has been argued that distributional semantic models that represent terms by a single vector cannot adequately handle polysemy, since they conflate several different usage patterns in one and the same vector \cite{Erk:2010,Veronis:2004}. Examples like the one above is often cited as evidence. We argue that this critique is unfounded and misinformed, and that it is {\em the mode of querying} the distributional semantic model that can be susceptible to problems with polysemy. As the above example demonstrates, querying distributional semantic models by $k$-NN conflates different usages of terms. The reason for this seems quite obvious: simply ranking the nearest neighbors by similarity (or distance) ignores any local structures of the neighborhood. If "suit" has as neighbors both "dress" and "lawsuit", which represent two distinct types of usages of "suit", there will be a {\em structural} distinction in the neighborhood of "suit" between these different neighbors, since they will be mutually unrelated (i.e.~there is a similarity between "suit" and "dress" and between "suit" and "lawsuit", but {\em not} between "dress" and "lawsuit").

$k$-NN also gives rise to another problem related to polysemy in distributional semantic models. The problem is that the most frequent senses
%% vi bör bestämma oss för usages eller senses
will populate the top of the nearest neighbor list, while the less frequent senses will not appear until further down the list, and if we set a too restrictive $k$, we will only see neighbors relating to the most frequent sense. Consider, for example, a term such as "suit", which, as we have seen above, may appear in (at least) two different senses: in usages related to \emph{law} and in usages related to \emph{clothes} (or {\em garment}). The distributional vector can be thought of as a sum $v_{suit} = f_{suit|law}v_{suit|law} + f_{suit|clothes}v_{suit|clothes}$, where $v_{suit|law}$ is an idealized notion of the {\em true} distributional vector of "suit" in the \emph{law}-sense, and $f_{suit|law}$ is the relative frequency of this sense.\footnote{Weighting schemes muddles this notion quite a bit, but we think the general intuition still holds.} From there one can easily argue that a similarity such as $s(v_{suit}, v_{garment})$ is actually a weighted composite of the similarities $s(v_{suit|law}, v_{garment})$ and $s(v_{suit|clothes}, v_{garment})$.\footnote{In the case of cosine similarity this follows nicely from the distributive property of dot products: $v=av_1+bv_2$, $s(v,w) = \frac{v\cdot w}{\|v\|\|w\|}=\frac{a(v_1\cdot w) + b(v_2\cdot w)}{\|v\|\|w\|}$} If "suit" occurs predominantly in the \emph{law}-sense in our corpus, the $k$-NN neighborhood of "suit" will be dominated by words pertaining to its \emph{law}-sense, while the less frequent senses might not be present at all. A misguided $k$ may thus obscure any other, less frequent, senses of a term.

\begin{figure}[h]
 	\centering
	\includegraphics[scale=0.31]{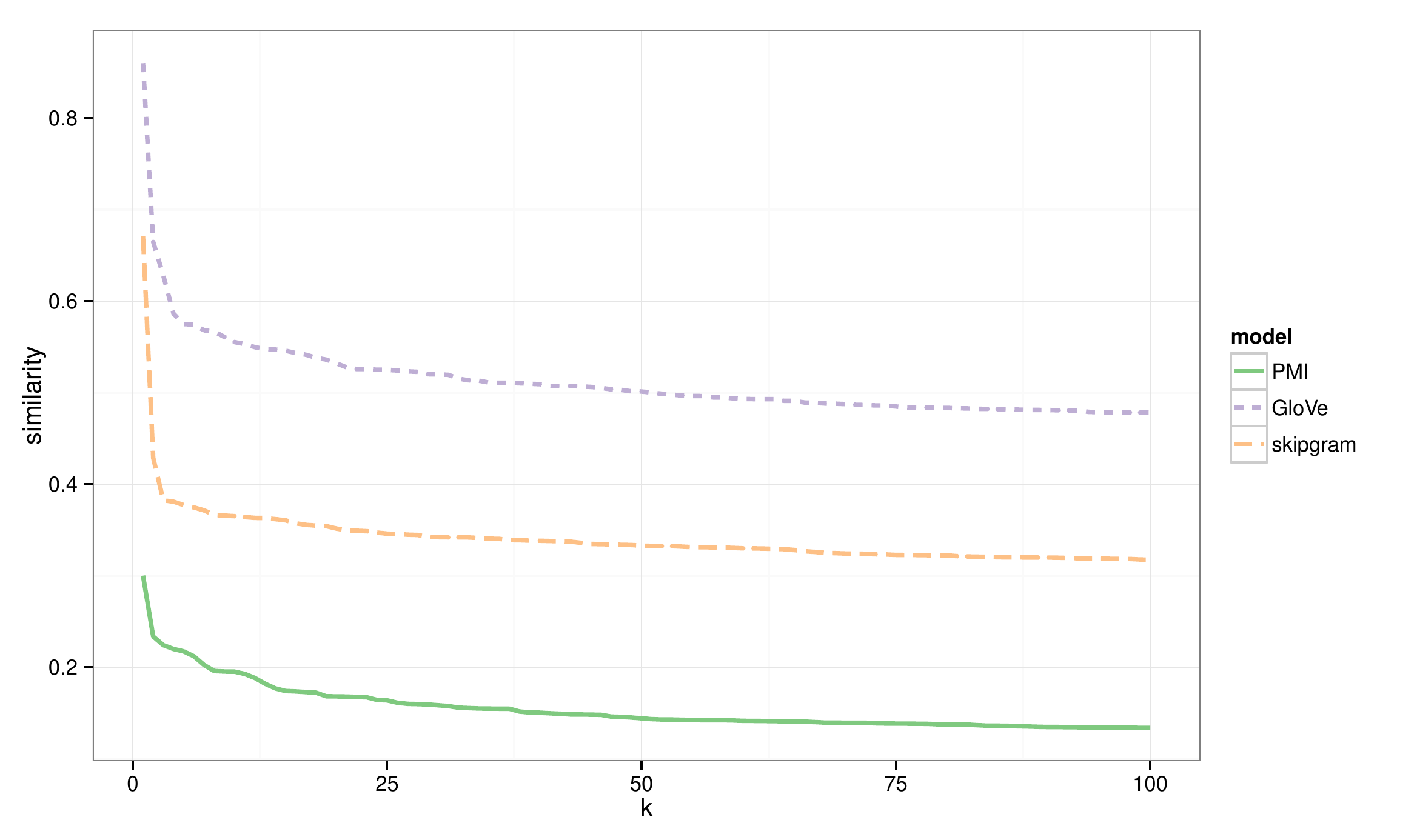}
	\caption{Distribution of similarities for the 100 nearest neighbors of the word \emph{suit} in the three distributional semantic models used in this paper.}
	\label{fig:dist}
\end{figure}

Another problem with setting a global $k$ in distributional semantic models is that some terms will have a much denser neighborhood than others. Using the same $k$ for all terms therefore seems ill-advised; terms with a dense neighborhood warrant a larger $k$ than those with a sparse neighborhood. As we have already touched upon in the introduction, determining $k$ is a fundamental question in $k$-NN, for which there seems to be no clear solution. A more informed approach compared to setting a global $k$ would be to consider the distribution of distances/similarities and attempt to find a gap in the distribution at which to cut off the list. However, the distribution of similarities in distributional semantic models 
typically does not have any clear gaps,
%is typically continuous and without any clear gaps,
% * <magnus.sahlgren@gavagai.se> 2014-12-30T08:58:09.216Z:
%
%  bättre term än "gaps"?
%
as exemplified in Figure \ref{fig:dist}. 

\begin{figure}[h]
\centering
\includegraphics[scale=0.31]{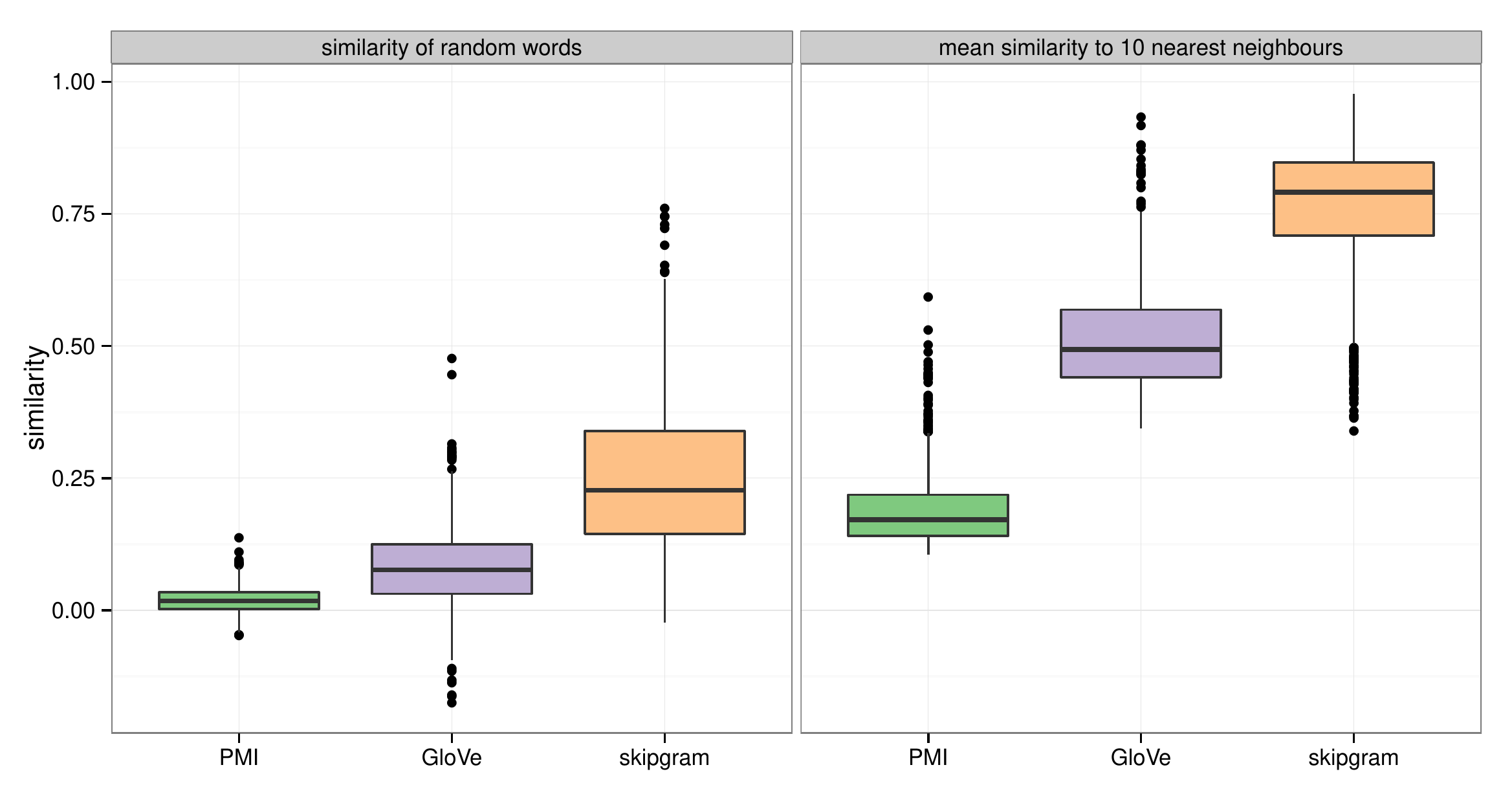}
\caption{Boxplots of the similarities of 1000 randomly picked word pairs (right), and of the mean similarities to the 10 nearest neighbors for 1000 randomly chosen words (left) in the three distributional semantic models used in this paper.}
\label{fig:box}
\end{figure}

Note that the curves behave approximately the same in all three models; there are a few (one or two) very close neighbors, and then the similarities decrease very slowly. The difference in magnitude of the similarities between the models is not peculiar for the word "suit". On the contrary, PMI, GloVe, and the skipgam model produce vector spaces with different inherent densities. Figure \ref{fig:box} shows both the similarities to 1000 randomly selected points, as well as the similarities to the 10 nearest neighbors to 1000 randomly selected points. The skipgram model produces the highest similarity scores both for related and unrelated points, while the PMI model produces the lowest scores for both related and unrelated points. The GloVe model is in between. All models show a more or less clear distinction between the average similarities to randomly chosen points and the average similarities to the nearest neighbors. This distinction suggest that it might be possible to use the expected similarity to a randomly selected point as a cut-off threshold for $k$-NN. However, such a global estimate will not be suitable for all terms, for the very same reason alluded to above; different terms have different densities of their neighborhoods. Furthermore, it seems as if the PMI model distinguishes more clearly between the related and the unrelated points, with the skipgram model having the most outliers. This suggests a global estimate might be more useful in some types of models (like the standard PMI model) than in others (like the skipgram model). 
% We use the term {\em horizon} to refer to a definition of $k$ that includes all relevant neighbors for a given term. We suggest that relative neighborhood graphs presents a (partial) solution to the problem of finding a relevant $k$ for a given term.

\section{Word-sense Induction}
\label{sec:wsi}

Selecting a relevant $k$ for a given term and grouping the neighbors according to which senses they represent is an example of {\em word-sense induction}. Distributional semantic models are well suited for this task, and there have been a number of different approaches suggested in the literature, which can roughly be divided into {\em context clustering} and {\em word clustering} approaches.
%% beroende på sty-fil får vi ändra alla cite i subjektsposition innan submission
Context clustering does not operate on nearest neighbor lists, but instead clusters representations of each individual occurrence of a term. \cite{Schutze:1998} is one of the earliest examples of a context clustering approach, in which {\em context vectors} (the centroid of the distributional vectors of the terms that occur in the context) for a given term are clustered into a set of {\em sense vectors} that represent the induced senses. Other examples of context clustering include \cite{Purandare:2004}, \cite{Velldal:2005}, \cite{Reisinger:2010}, \cite{Pedersen:2010}, and \cite{Jurgens:2010}.

In contrast to context clustering
%approaches
, word clustering 
%approaches
clusters the nearest neighbors into sense groups, and are thus the type of approach that is most relevant for our purposes. The earliest example of a word clustering approach is {\em distributional clustering} \cite{Pereira:1993}, which clusters nouns 
%($n$)
that occur as heads of direct objects of verbs 
%($v$) 
according to their distributional similarity. The resulting noun clusters 
%($c$)
for a verb can be interpreted as a representation of the different senses of the verb. 
%Distributional clustering is based on a probabilistic decomposition model $\sum_{c \in C} p(c)p(n|c)p(v|c)$ that uses maximum likelihood estimation to fit the model to observed data.

Another example of a word clustering approach is {\em clustering by committee} \cite{Pantel:2002}, which is a distributional clustering procedure in several steps. The first step is to use average-link clustering to recursively cluster the nearest neighbors into a set of clusters called {\em committees}. The committees are then used to define clusters by iteratively adding committees whose similarity to the term exceeds a certain threshold, and that is not too similar to any other added committee. For each added committee, its features are also removed from the distributional representation of the lexeme. This last step ensures that the clusters do not become too similar, and that clusters representing less frequent senses can be discovered. 

The idea of iteratively removing features from the distributional vector when a sense cluster as been formed is also present in \cite{Dorow:2003}, who use a graph-based clustering method (Markov clustering \cite{Dongen:2000}) to cluster the nearest distributional neighbors of a lexeme. Another graph-based approach to word-sense induction is the {\em HyperLex} algorithm \cite{Veronis:2004}, which constructs a graph connecting all pairs of terms that co-occur in the context of an ambiguous term. The resulting graph contains highly connected components (hubs), which represent the different senses of the term. \cite{Agirre:2006} compares HyperLex to {\em PageRank} \cite{Brin:Page:1998} and demonstrates that the two methods perform similarly on a word-sense induction task. Other examples of graph-based approaches include \cite{Biemann:2006}, \cite{Klapaftis:2008}, and \cite{diMarco:2013}.

There have also been several attempts to use various types of matrix factorization to perform word sense induction. The idea is that the factorization uncovers a set of global senses in the form of the latent factors, and that the sense distribution for a given term can be described as a distribution over these latent factors. \cite{Brody:2009}, \cite{Seaghdha:2011}, \cite{Yao:2011}, and \cite{Lau:2012} use different versions of {\em Latent Dirichlet Allocation} to produce the factorization, while \cite{Dinu:2010} and \cite{VandeCruys:2011} instead experiment with {\em non-negative matrix factorization}.

\cite{Tomuro:2007} argues that clustering approaches like distributional clustering or clustering by committee may produce clusters that are themselves polysemous, which may not be a desirable property of a word sense induction algorithm. As a solution to this problem, Tomuro et al.~suggest using {\em feature domain similarity}, which refers to the similarity between the {\em features} of items rather than the similarity between the items themselves. 
%The feature domain similarity between two distributional vectors $x$ and $y$ is defined as $\alpha \times \cos(x,y) + \beta \times \text{dt}(x+y)$, where $\alpha=0.95$ and $\beta=0.05$, $\cos(x,y)$ is the cosine similarity between $x$ and $y$, and $\text{dt}(x+y)$ is the {\em domain tightness} of $x+y$, given by $\frac{1}{n} \sum_{a,b \in F;a \neq b} \frac{z(a) + z(b)}{2}\text{wnSim}(a,b)$, where $F$ is the set of features of $z$, $n$ is the number of pairwise combinations of the features in $F$, and $\text{wnSim}(a,b)$ is the maximum similarity between the synsets of $a$ and $b$ in WordNet. 
%Tomuro et al.~incorporate 
The domain feature similarity score is incorporated in a modified version of the clustering by committee algorithm, in which the algorithm is run twice, using the output of the first run as input to the second run. The idea is that this iterative approach may enable the algorithm to utilize higher-order features,
%~\cite{Tomuro:2007}. The incorporation of domain feature similarity 
and that this will inhibit the formation of polysemous clusters, since the domain feature similarity of a polysemous cluster will be lower than the score for a monosemous cluster.

\cite{Tamm:2014} also leverage on the idea of using feature similarity as the basis of sense clustering. The approach, called {\em syntagmatically labeled partitioning}, relies on a distributional semantic model that encodes sequential as well as substitutable relations. The method essentially sorts the $k$ nearest (substitutable) neighbors according to which sequential connections they share. The resulting partitioning of the nearest distributional neighbors does not only constitute a word-sense induction, but it also provides {\em labels} for the induced senses in the form of the sequential connections the neighbors share. 
%% We will revisit this idea in the context of neighborhood graphs, in the following section.

\section{Neighborhood Graphs}
\label{sec:rng}

Many of the previous approaches to word-sense induction mentioned in the previous section operate at a global level, utilizing global structural properties of the semantic spaces, e.g.~by matrix factorization techniques. We believe this is as ill-advised as setting a global $k$ or radius for the nearest neighbor search, since it is the {\em local} structures that are important when analyzing nearest neighbors. Other approaches to word-sense induction use various forms of clustering techniques. However, 
%as \cite{Karlgren:2014} point out, the notion of a cluster is less useful in high dimensional spaces \cite{Beyer:1999}, since the transitivity of distance is not as obvious in high dimensions; if two points $a$ and $c$ are both close to another point $b$, they may still be at considerable distance from each other.
% * <magnus.sahlgren@gavagai.se> 2014-12-30T09:23:09.566Z:
%
%  lite klipp o klistrat - du bör nog skriva om meningen om "however, the notion..."
%
%Furthermore, 
previous studies of the intrinsic dimensionality of distributional semantic spaces using fractal dimensions indicate that neighborhoods in semantic space have a {\em filamentary} rather than clustered structure \cite{Karlgren:2008}.

We therefore propose the use of {\em topological} models that take the {\em local} structure of neighborhoods in semantic space into account. The method proposed here performs no global clustering, does not concern itself with grammatical preprocessing or parsing, and the distributional vectors are taken as is. The approach discovers different word senses from the local structure of neighborhoods, given nothing but similarities between points. As such it is easy to test on widely different vector models, as long as there exists a well behaved similarity function. The proposed approach not only answers the question which other terms are similar to a given term, but also {\em how} are they similar.

%% och med labelled rng kommer vi besvara frågan "why"

\begin{figure}[h]
\begin{subfigure}{.25\textwidth}
	\centering
	\includegraphics[scale=0.3,frame]{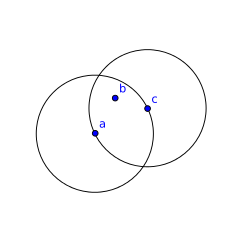}
%%\caption{A point $b$ that is between points $a$ and $c$.}
%%\label{fig:inside}
\end{subfigure}
\begin{subfigure}{.2\textwidth}
	\centering
	\includegraphics[scale=0.3,frame]{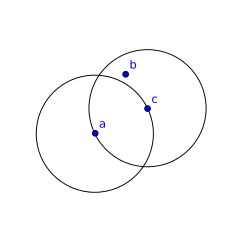}
%%	\caption{A point $b$ that is not between points $a$ and $c$.}
%%	\label{fig:outside}
    \end{subfigure}
    \caption{Example of when point $b$ is between point $a$ and $c$ (left), and when it is not (right).}
    \label{fig:rng}
\end{figure}

{\em Relative} neighborhoods are examples of {\em empty region graphs} \cite{Cardinal:2009}, where points are neighbors if some region between them is empty. 
%For relative neighborhoods, the neighborhood of a point $a$ in some set of points $V$ are all points of $V$ that do not have any other point "between" them and $a$. 
%Alternatively: two points $a,c \in V$ are neighbors iff the region $U(a,c)$ is empty. 
%{\em Relative} neighborhood graphs (RNG) are one example of empty region graphs, and the only example we will cover here. 
For relative neighborhood graphs the region between two points $a$ and $c$ belonging to some set of points $V$ is defined as the intersection of the two spheres with centers in $a$ and $c$, with radius $d(a,c)$. In other words, a point $b$ lies between points $a$ and $c$ if it is closer to both $a$ and $c$ than $a$ and $c$ are to each other, and if no such point $b$ exists, $a$ and $c$ are neighbors. 
%Ergo: if $d(a,b) < d(a,c)$ and $d(b,c) < (a,c)$ then $b$ lies between $a$ and $c$, and if no such $b$ exists --- if the region between $a$ and $c$ is empty --- then $a$ and $c$ are said to be neighbors. 
Illustrations of this can be seen in Figure \ref{fig:rng}. 

Such neighborhoods have been argued to better preserve local topology \cite{Bremer:2014}, and be more robust to deformations of the data than $k$-NN neighborhoods \cite{correa:2012} as they in some sense contain information about direction whereas $k$-NN neighborhoods only contain information about distance. Going back to the "suit" example, we can see that if "suit" in the law sense is more similar to the composite "suit" than to its clothes sense, and vice versa, then the composite $v_{suit}$ lies between  $v_{suit|law}$ and $v_{suit|clothes}$. This in turn means that out of those two points, both are relative neighbors to "suit", and neither of them lies between the other and "suit". 

% * <magnus.sahlgren@gavagai.se> 2014-12-18T15:50:52.345Z:
%
%  går det att argumentera lite mer kring detta här? vore kanoners!
%
% ^ <magnus.sahlgren@gavagai.se> 2015-01-07T12:15:58.293Z.
Formally, the set of points between two points $a,c \in V$ can be characterized and computed in the following way: 
\[\text{betweens}(V,a,c) = \{b|b\in V,b \text{ lies between } a \text{ and } c\}\]
\[\text{rng-nbh}(V,a) = \{c|c \in V, \text{betweens}(V,a,c)=\emptyset \}\]
\[E_\text{rng}(V) = \{(a,b) | a \in V, b \in \text{rng-nbh}(V,a)\}\]
\noindent
where $E_\text{rng}$ is the undirected edge set of the RNG. The function $\text{betweens}(V,a,c)$ can be straightforwardly translated to an algorithm taking  $\mathcal{O}(|V|)$ time, making the rng-nbh function take $\mathcal{O}(|V|^2)$ time, which in turn makes the computation of the complete graph take $\mathcal{O}(|V|^3)$ time.\footnote{Assuming a constant time distance function.} Clearly unfeasible, but we have not found any alternatives that performs better in the high dimensional case.\footnote{It should be noted that there are more efficient algorithms for lower dimensional situations.}  
%Algorithm \ref{alg:rng} presents non-declarative flavors of the above. 

%\begin{algorithm}
%\caption{Between}
%\label{alg:between}
%\begin{algorithmic}[]
%	\Function{between}{$V, a, c$}
%    	\State $bs \gets \emptyset$ 
%		\ForAll{$b \in V$}
%            \If{$d(a,c) > d(a,b) \land d(a,c) > d(b,c)$}
%            	\State add $b$ to $bs$ 
%            \EndIf
%      	\EndFor
%  		\State \Return $bs$ 
%	\EndFunction 
%\end{algorithmic}
%\end{algorithm}

%\begin{algorithm}
%\caption{Relative Neighborhoods}
%\label{alg:rng}
%\begin{algorithmic}[]
%	\Function{between}{$V, a, c$}
%    	\State $bs \gets \emptyset$ 
%		\ForAll{$b \in V$}
%            \If{$d(a,c) > d(a,b) \land d(a,c) > d(b,c)$}
%            	\State add $b$ to $bs$ 
%            \EndIf
%      	\EndFor
%  		\State \Return $bs$ 
%	\EndFunction 
%    \State
%	\Function{rng-nbh}{$V, a$}
%    	\State $nbh \gets \emptyset$
%		\ForAll{$c \in V$}
%        	\If{\Call{between}{$V,a,c$}=$\emptyset$}
%            	\State add $c$ to $nbh$
%            \EndIf
%         \EndFor
%		\State \Return $nbh$ 
%	\EndFunction
%    \State
%    \Function{E-rng}{$V$}
%    	\State $E \gets \emptyset$ 
%        \ForAll{$a \in V$}
%        	\ForAll{$c \in~$\Call{rng-nbh}{$V,a$}}
%            	\State add $(a,c)$  to $E$ 
%            \EndFor
%        \EndFor
%        \State \Return $E$
%	\EndFunction
%\end{algorithmic}
%\end{algorithm}
% * <magnus.sahlgren@gavagai.se> 2014-12-30T11:50:06.176Z:
%
%  och algoritm 3?
%
% ^ <amarucuba@gmail.com> 2015-01-07T08:51:41.371Z.

In \cite{correa:2012} it is noted that the intersection of the relative neighborhood graph and the $k$-NN graph is a more feasible alternative: 
\[\text{k-rng-nbh}(V,a) = \text{rng-nbh}(V',a)\]
\[\text{where}~V' = k\text{ nearest neighbors of } a\]
Given a  precompiled --- i.e. constant time --- k-NN lookup,  the above takes $\mathcal{O}(k^2)$ time, so using a heap-based $\mathcal{O}(|V|\lg k)$ $k$-NN algorithm results in an algorithm  taking $\mathcal{O}(k^2 + |V|\lg k)$ time.
%which makes the complete computation of $E_\text{k-rng}(V)= \{(a,b)|a \in V, b \in \text{k-rng-nbh}(V,a)\}$ take $\mathcal{O}(|V|(k^2 + |V|\lg k))$ time using a heap-based  $\mathcal{O}(|V|\lg k)$ $k$-NN algorithm.  Algorithm \ref{alg:k-rng}.

%\begin{algorithm}
%\caption{k-RNG}
%\label{alg:k-rng}
%\begin{algorithmic}[]
%	\Function{k-rng-nbh}{V,a}
%    	\State $V' \gets$ \Call{k-NN}{V,a}
%        \State \Return \Call{rng-nbh}{$V',a$}
%    \EndFunction
%    \Function{E-k-rng}{$V$}
%    	\State $E \gets \emptyset$ 
%        \ForAll{$a \in V$}
%        	\ForAll{$c \in~$\Call{k-rng-nbh}{$V,a$}}
%            	\State add $(a,c)$  to $E$ 
%            \EndFor
%        \EndFor
%        \State \Return $E$
%	\EndFunction
%\end{algorithmic}
%\end{algorithm}

The same idea can be used to build a tree structure --- here called {\em relative neighborhood tree} --- rooted in a reference word $a$ , in the following way: 
\[\text{rnbh-tree}(V,a) = \{(c,\argmin_{b\in B_c }d(b,c))|c \in V\}\]
\[\text{where}~B_c = \{a\}\cup \text{betweens}(V,a,c)\]

%\[\text{rnbh-tree}(V,a) = \{(c,\argmin_{b\in bs\cup \{a\} }d(b,c))|c \in V, bs =\text{between}(V,a,c)\}\]
\noindent 
%or as described in Algorithm \ref{alg:rnbh-tree}. 
This can easily be restricted to the k-nearest neighbors of $a$ in much the same way as above. 

%\begin{algorithm}
%\caption{Relative neighborhood tree}
%\label{alg:rnbh-tree}
%\begin{algorithmic}[]
%	\Function{rnbh-tree}{V,a}
%    	\State $E \gets \emptyset$  
%        \ForAll{$c\in V$}
%        	\State $min \gets a $ 
%       		\ForAll{$b \in $ \Call{between}{V,a,c}}
%            	\If{$d(b,c) < d(min,c)$}
%                	\State $min \gets b$ 
%                \EndIf
%            \EndFor
%            \State add $(c,min)$ to $E$ 
%        \EndFor
%        \State \Return $E$ 
%    \EndFunction
%\end{algorithmic}
%\end{algorithm}

\begin{figure}[h!]
\centering
\includegraphics[scale=0.18]{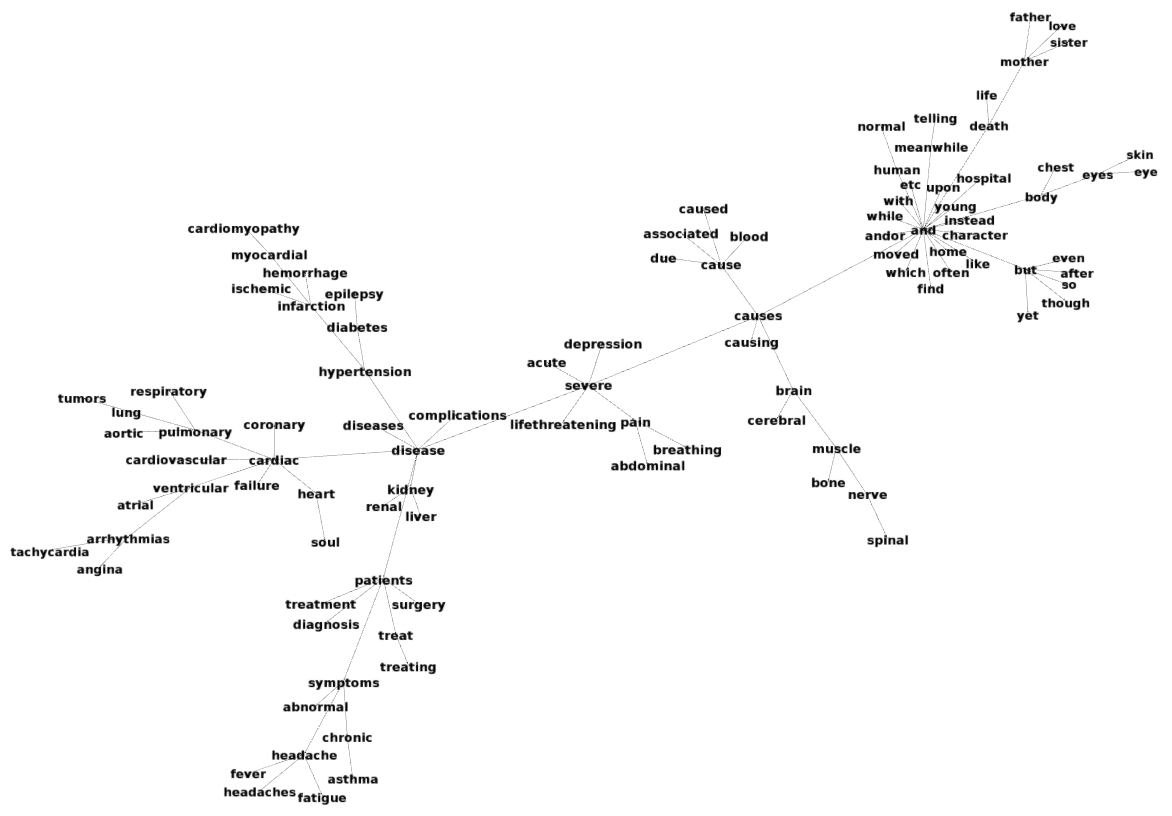}
\caption{Relative neighborhood tree for "heart" in the PMI model, restricted to the 100 closest neighbors.}
\label{fig:pmi}
\end{figure}

Computing this for a point $a$  produces a tree where the direct children of $a$ are its relative neighbors, and the parent of a point $c$  further down the tree is the point between $a$ and $c$ that is closest to $c$. Figure~\ref{fig:pmi} illustrates what the resulting structure looks like for "heart" on its 100 nearest neighbors in the PMI model. Note that the root "heart" (at the mid-left in the graph) only has two relative neighbors: "cardiac" and "soul." One advantage of using this type of structure for the neighborhood is that it enables us to examine various depths of the tree. Depth one includes only the direct neighbors ("cardiac" and "soul"); depth two includes all neighbors two steps away in the graph: "disease," "coronary," "pulmonary," "cardiovascular," "ventricular," and "failure," which are all children to "cardiac;" depth three also includes the neighbors "kidney," "severe," "complications," and "diseases" as children to "disease," "atrial" and "arrhythmias" as children to "ventricular," and the neighbors "respiratory," "lung," "tumors," "aortic" as children to "pulmonary." This tree structure can be used to identify neighbors that are themselves polysemous (c.f.~the critique mentioned in Section~\ref{sec:wsi} of clustering-based approaches to word-sense induction that they may produce polysemous clusters \cite{Tomuro:2007}). One example is the neighbor "disease" at depth two, which has six children that refer to different aspects of disease. %Figure \ref{fig:pmi20}
%to \ref{fig:skipgram20} 
%shows the same tree restricted to the 20 nearest neighbors for the PMI model.
%all models. 
%By taking into account the depth of the RNG, we can see that although certain terms are among the $k$ nearest neighbors, they can be far away in the graph.

This tree-structure is thus quite useful in the context of word-sense induction, since the branching structure indicates different usages, and the depth factor enables us to calibrate the granularity of the induced word senses. If we only consider direct neighbors (i.e.~depth one), and set $k=V$ (i.e.~we do an exhaustive nearest neighbor search), we will extract all terms that have a direct connection to the reference term. We refer to this neighborhood as the {\em semantic horizon}. At the most coarse level of analysis, this is the neighborhood that represents the main induced senses of a term. Tables \ref{tab:rng-suit} and \ref{tab:rng-bad} provide examples of $1000$-RNG neighborhoods. 

% * <magnus.sahlgren@gavagai.se> 2014-12-20T18:21:30.463Z:
%
%  utveckla och exemplifiera!
%
% ^ <magnus.sahlgren@gavagai.se> 2015-01-02T15:49:06.926Z.

%% det vore intressant att lägga in en avgränsare i tabellerna där likheterna blir mindre än det förväntade slumpavståndet
%% funderat på det: är det medelavståndet man vill åt eller medianavståndet? troligtvis det senare ... å medianavståndet är definitionsmässigt inte representerat (eftersom det finns >2000 ord) 
%% det vore även cool att kunna visa en grafisk representation av hur rng-grannskapen ser ut - använda gephi, typ

%\begin{table}[h]
%\centering
%\caption{Relative neigbhourhood of the word ``suit'' in a number of different DSMs.}
%\label{tab:rng-suit}
%\begin{tabular}{|l|l|l|}
%PMI & GloVe & skipgram \\
%\hline
%suits (1)&suits (1)&suits (1)\\
%dress (2)&lawsuit (2)&lawsuit (2)\\
%lawsuit (10)&mobile (33)&\\
%dinosaur (53)&gundam (34)&\\
%costly (60)&trump (55)&\\
%option (76)&zoot (133)&\\
%counterparts (99)&rebid (423)&\\
%predator (107)&serenaders (458)&\\
%trump (109)&hev (987)&\\
%$\vdots$&&\\
%\end{tabular}
%\end{table}

\begin{table}[ht!]
\scriptsize{
\centering
\caption{Relative neighborhood of the words ``suit,'' "orange," and "heart" in three different semantic models. The numbers in parenthesis indicate the $k$-NN ranks of the neighbors.}
\label{tab:rng-suit}
\begin{tabular}{|l|l|l|}
\hline
{\bf PMI} & {\bf GloVe} & {\bf skipgram} \\
\hline
\multicolumn{3}{|c|}{suit} \\
\hline
suits (1)&suits (1)&suits (1)\\
dress (2)&lawsuit (2)&lawsuit (2)\\
lawsuit (10)&mobile (33)&\\
dinosaur (53)&gundam (34)&\\
costly (60)&trump (55)&\\
option (76)&zoot (133)&\\
counterparts (99)&rebid (423)&\\
predator (107)&serenaders (458)&\\
trump (109)&hev (987)&\\
$\vdots$&&\\
\hline
\multicolumn{3}{|c|}{orange} \\
\hline
yellow (1)&yellow (1)&redorange (1)\\
lemon (16)&ktype (12)&\\
&lemon (14)&\\
&citrus (17)&\\
&jersey (21)&\\
&cherry (24)&\\
&county (26)&\\
&peel (42)&\\
&jumpsuits (57)&\\
&$\vdots$&\\
\hline
\multicolumn{3}{|c|}{heart} \\
\hline
cardiac (1)&my (1)&congestive (1)\\
soul (22)&blood (2)&hearts (2)\\
hearts (183)&throbs (3)&\\
ashtray(641)&suffering (4)&\\
rags(771)&brain (6)&\\
&cardiac (8)&\\
&hearts (11)&\\
&throb (17)&\\
&lungs (22)&\\
&$\vdots$&\\
\hline
\end{tabular}
}
\end{table}

These examples demonstrate some interesting similarities and differences between the three models. First of all, there are some direct neighbors that are present in all three models: "suit" has "suits" and "lawsuit" as direct neighbors in all three models, "heart" has "hearts," "service" has "services," and "above" has "below". Plural forms are of course reasonable neighbors of their singular counterparts in a semantic model, but their usefulness for word-sense induction can perhaps be questioned. Taking "suits" to indicate the {\em garment}-related sense of "suit," all three models produce both a {\em garment}-related and {\em law}-related sense. For "orange," the skipgram model only represents the {\em color} sense, while the PMI and GloVe models also feature a {\em fruit} sense. For "heart," all three models have a {\em disease} sense (represented by the neighbors "cardiac" in the PMI and GloVe models, and the neighbor "congestive" in the skipgram model), and an {\em organ} sense (represented by the plural form "hearts"). "Service" is a comparably vague term that has a number of different senses in the PMI and GloVe models, nut only one in the skipgram model. "Bad" produces both a {\em negativity} sense and a {\em German spa town}-sense in all three models, both only the GloVe and skipgram models have a separate antonym sense ("good" is not a direct neighbor in the PMI model). "Above" has both the antonym and direct neighbors relating to measurements in all three models.

GloVe produces the most branched neighborhoods, with a large number of direct neighbors, while the skipgram model produces the least branched neighborhoods with at most a couple of direct neighbors for each term. The PMI model is somewhere in between. One reason why GloVe produces such branching neighborhoods is that GloVe seems to capture not only semantic relations but also a significant amount of sequential relations. Many of the neighbors in the $k$-RNG for GloVe come from collocations: the $k$-RNG for "suit" includes "mobile" and "gundam," which come from the collocation "mobile suit gundam" that is an anime series, "trump" that relates to "trump suits" in card games, "serenaders" that refer to the retro string band "cheap suit serenaders," and the very distant neighbor "hev," which comes from "hev suit" that relates to the Half-life series of first person shooter games. For "orange" we find "ktype" that comes from the collocation "ktype stars," which is another term for "orange dwarfs", as well as the collocations "orange peel", "orange county", "orange jumpsuit", "cherry orange", and so on. The $k$-RNG for "heart," "service," "bad," and "above" also feature a number of collocations for the GloVe model. There are also some examples of neighbors from sequential relations in the PMI model (e.g.~"costly" as neighbor to "suit" from the collocation "costly suit," "luck" and "donnersbergkreis" as neighbors to "bad" from the collocations "bad luck" and "bad donnersbergkreis"), but this tendency is not at all as pronounced as it is for the GloVe model.

\begin{table}[ht!]
\scriptsize{
\centering
\caption{Relative neigbhors to the word "service," ``bad,'' and "above" in three different semantic models. The numbers in parenthesis indicate the $k$-NN ranks of the neighbors.}
\label{tab:rng-bad}
\begin{tabular}{|l|l|l|}
\hline
{\bf PMI} & {\bf GloVe} & {\bf skipgram} \\
\hline
\multicolumn{3}{|c|}{service}\\
\hline
services (1)&services (1)&services (1)\\
network (2)&operated (3)&\\
operates (8)&serving (6)&\\
launched (18)&military (17)&\\
served (22)&duty (20)&\\
intercity(34)&passenger (21)&\\
&dialaride (644)&\\
&aftersales (759)&\\
&limitedstop (802)&\\
\hline
\multicolumn{3}{|c|}{bad}\\
\hline
terrible (1)&good (1)&nauheim (1)\\
that (2)&kissingen (2)&good (2)\\
luck (39)&ugly (45)&dreadful (5)\\
unfortunate (70)&nasty (48)&\\
stalling (276)&dirty (106)&\\
donnersbergkreis (860)&omen (328)&\\
rancid (980)&conkers (360)&\\
&karma (952)&\\
\hline
\multicolumn{3}{|c|}{above}\\
\hline
below (1)&below (1)&below (1)\\
around (2)&level (2)&500ft (2)\\
feet (5)&height (3)&\\
measuring (29)&just (4)&\\
beneath (36)&stands (10)&\\
columns (62)&lower (11)&\\
atop (102)&beneath (12)&\\
&rise (21)&\\
&sea (30)&\\
&$\vdots$&\\
\hline
\end{tabular}
}
\end{table}

\begin{figure}[ht!]
\centering
\includegraphics[scale=0.19]{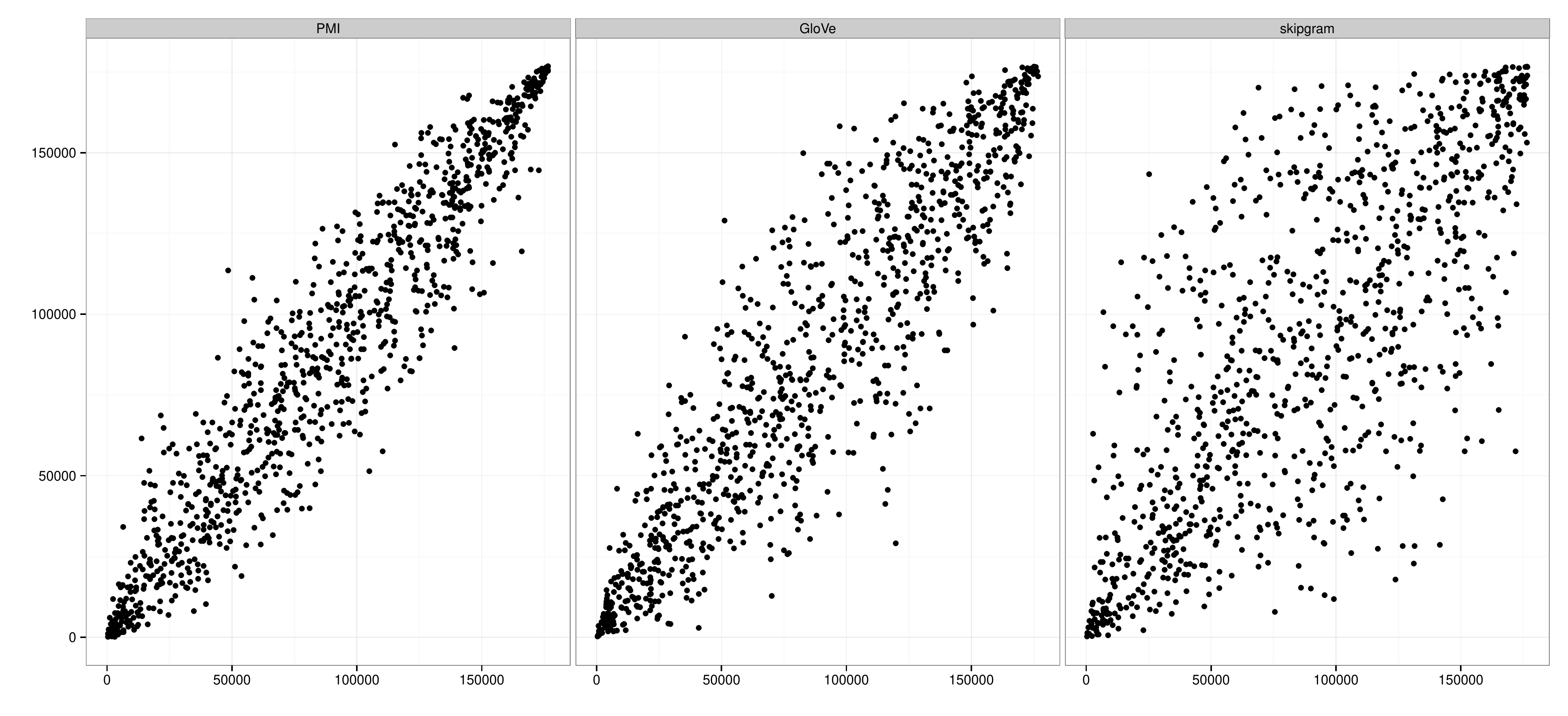}
\caption{neighborhood reciprocity in the different models.}
\label{fig:corrplot}
\end{figure}

The PMI and GloVe models produce the structurally most similar RNGs, with on average a handful of direct neighbors, of which some can be very distant. The skipgram model on the other hand produces very few direct neighbors.
% * <magnus.sahlgren@gavagai.se> 2014-12-20T18:27:02.067Z:
%
%  tankar om varför det blir så för skipgram?
%
% ^ <amarucuba@gmail.com> 2015-01-05T14:51:19.932Z.
This led us to look further into the structural properties of neighborhoods in the skipgram model. An interesting observation --- and possible complication --- is that the neighborhoods in the skipgram model are highly asymmetric: the first neighbor of "information" is "informations", whereas "information" is only the 1829th neighbor of "informations." While such asymmetry occurs in all models, it seems much more prevalent in the skipgram model. Figure \ref{fig:corrplot} demonstrates this: each point corresponds to a random word pair $(a,b)$ with x corresponding to where $b$ is in the ordered list of $a$'s neighbor, and $y$ to where $a$ is in the ordered list of $b$'s neighbors, or equivalently: x is the number of points within $d(a,b)$ of $a$ and y is the number of points within $d(a,b)$ of $b$. This implies that the local densities vary much more in the skipgram model than in the others, which can complicate the choice of $k$ in the $k$-RNG algorithm.

\section{Conclusions}

In this paper we have discussed the question how to query semantic models, which is a question that has been long neglected in research on computational semantics. Nearest neighbor search (or $k$-NN) is often treated as the only available option, which leads to misunderstandings regarding how semantic models represent and handle vagueness and  polysemy. We have argued that the structure --- or topology --- of the local neighborhoods in semantic models carry useful semantic information regarding the different usages --- or senses -- of a term, and that such topological properties thus can be used to analyze polysemy and to perform word-sense induction. We have also argued that the topology of the local neighborhoods in semantic models can be used for selecting a relevant set of neighbors --- a factor we have referred to as the semantic horizon. 

We have introduced relative neighborhood graphs (RNG) as an alternative to standard $k$-NN, and we have illustrated how  $k$-RNG can be used as a tool for analyzing the topology of local neighborhoods in semantic models. We have exemplified relative neighborhoods in three different well-known semantic models; the standard PMI model, as well as the more recent GloVe and skipgram models. The examples provided in this paper demonstrate that $k$-RNG can be used for word-sense induction, but that such topological methods are more suitable to use for certain types of semantic models. The k-RNG for the PMI and GloVe models produce pleasant results, while the skipgram model, with its big local variations in density, produces less informative results. It's quite possible that the complete RNG overcomes these problems, but that does not seem a feasible solution.% * <magnus.sahlgren@gavagai.se> 2015-01-02T15:31:33.977Z:
%
%  that are...? vad? hur? varför?
%
% ^ <amarucuba@gmail.com> 2015-01-07T14:31:43.982Z.

This illustrates how $k$-RNG can be used as a tool to gain insight into the topological properties of different models. We have also observed that the GloVe model often produces neighbors that correspond to various collocations, which means that this model is not strictly a {\em semantic} representation, since it confounds substitutable and sequential relations. A more sophisticated tokenization, taking n-grams into account, might alleviate this. The standard PMI model is nowadays often overlooked in favor of more recent neural network-inspired models, but our results indicate that the PMI model has a number of comparatively attractive properties that are useful for linguistic applications such as word-sense induction.

%To conclude this paper, we submit the proposition that relative neighborhood graphs present an attractive alternative to standard $k$-NN, since it preserves the local topology of neighborhoods in semantic models.

\bibliographystyle{acl}
\bibliography{rng.bib}

\end{document}